\documentclass[conference]{IEEEtran}
\ifCLASSINFOpdf
\else
\fi
\newtheorem{thm}{{\textbf{Theorem}}}
\newtheorem{defn}{{\textbf{Definition}}}

\newtheorem{lem}{{\textbf{Lemma}}}

\begin{document}
\title{A Forgetting-based Approach to Merging Knowledge Bases}
\author{\IEEEauthorblockN{Dai Xu, Xiaowang Zhang and Zuoquan Lin}
\IEEEauthorblockN{Department of Information Science, Peking University\\
Beijing 100871, P. R. China\\
Email: \{xudai,zxw,lzq\}@is.pku.edu.cn }}

\maketitle \noindent\begin{abstract}
%
This paper presents a novel approach based on variable forgetting,
which is a useful tool in resolving contradictory by filtering some
given variables, to merging multiple knowledge bases. This paper
first builds a relationship between belief merging and variable
forgetting by using dilation. Variable forgetting is applied to
capture belief merging operation. Finally, some new merging
operators are developed by modifying candidate variables to amend
the shortage of traditional merging operators. Different from model
selection of traditional merging operators, as an alternative
approach, variable selection in those new operators could provide
intuitive information about an atom variable among whole knowledge
bases.
\end{abstract}


\IEEEpeerreviewmaketitle

\section{Introduction}
The knowledge bases (KBs) contain a large amount of information
coming from different sources. KBs must be able to ``{\em
intelligently}'' manage such distributed information. An important
task is ensuring that those KBs, comprising collections of
information which possibly conflict with each other, need to be
combined into a consistent whole (\cite{liberatore1998arbitration}).
As one of traditional ways of management, belief merging concerns
with the problem of determining a group's beliefs from individual
members' beliefs
(\cite{konieczny2002merging,gorogiannis2008merging}).
The merging process from the point of logic is formalize as follows:
let $\Phi=\{\varphi_1,\cdots,\varphi_n\}$ be a group of KBs (which
is composed of multiple formulae and can also be taken as
conjunction of them from logic equivalence) where each logical
formula $\varphi_i$ denotes a knowledge base(KB) from some
information source. $\Phi$ is a multi-set of formulae which is to
merge. $\mu$ is a set of formulae which represents the integrity
constraints, i.e., some information that the result of merging must
obey. The goal of merging is to obtain a new KB
$\triangle_{\mu}(\Phi)$ which represents the consensus of $n$
sources given integrity constraints $\mu$ (\cite{konieczny2004}).

When conflicts occur in KBs, we intuitively weaken them to achieve
consistency again. Thus we can avoid the trivialization of inference
that everything can be deducted from inconsistent KBs. Of course, a
fundamental requirement is to minimize changes of the original KBs
in maintaining their consistency. That is, the information of the
original KBs should be preserved as much as possible. Those current
approaches are based on two fundamentally different standpoints: the
syntax-based and the semantics-based. Based on syntax, the maximal
consistent subsets of the original bases consistent with the
integrity are selected as the merged result. Unfortunately, the
common weakness is that the merged results depend on the syntax
forms of KBs. Based on semantics (models), those models which have
the minimal distance to models of the KBs are selected from the
models of integrity constraints as the candidates models of the
merged result. The Dalal distance has been proved a useful way to
characterize two models (\cite{dalal1988investigations}).
Based on three methods of aggregating distances of multi-KBs, there
are three merging operators, namely, $\triangle_{\mu}^{\Sigma}$,
$\triangle_{\mu}^{Max}$ and $\triangle_{\mu}^{GMax}$.
$\triangle_{\mu}^{\Sigma}$ is taking the summation of distances as
the aggregation to pick out the most popular models
(\cite{lin1999knowledge}). $\triangle_{\mu}^{Max}$ is taking the
maximum of distances as the aggregation to minimize the worst cases
(\cite{revesz1993semantics}). $\triangle_{\mu}^{GMax}$ is based on
lexicographical order to capture the arbitration behavior of
operator $\triangle_{\mu}^{Max}$ (\cite{konieczny2002merging}).

However, three existing merging operators do not always work well in
some cases. We still consider an example of {\em swimming-pool}
discussed in \cite{konieczny2002merging} previously.

\noindent \textbf{Swimming-pool} At a meeting of a block of flat
co-owners, the chairman proposes for the coming year the
construction of a swimming-pool, of a tennis-court and of a
private-car-park. But if two of these three items are built, the
rent will significantly increase. We will denote by $S,T,P$
respectively the construction of the swimming-pool, the tennis-court
and the private-car-park. We denote $I$ the rent increase.
The chairman outlines that building two items or more will have an
important impact on the rent: $\mu=((S\wedge T)\vee (S\wedge P)\vee
(T\wedge P))\rightarrow I$.

There are four co-owners
$\Phi=\{\varphi_1,\varphi_2,\varphi_3,\varphi_4\}$. Two of the
co-owners want to build the three items and do not care about the
rent increase: $\varphi_1=\varphi_2=S\wedge T\wedge P$. The third
one thinks that building an item will cause at some time an increase
of the rent and wants to pay the lowest rent, So he is opposed to
any construction: $\varphi_3=\neg S\wedge \neg T\wedge \neg P \wedge
\neg I$. The last one thinks that the block really needs a
tennis-court and a private-car-park but does not want a high rent
increase: $\varphi_4=T\wedge P\wedge \neg I$. In
\cite{konieczny2002merging}, the merging results of applying
three operators above are in the following.\\
 $\triangle_{\mu}^{\Sigma}(\Phi)=S\wedge T\wedge
P\wedge I$.\\
 $\triangle_{\mu}^{Max}=(\neg S\wedge \neg T\wedge P)\vee
(\neg S\wedge T\neg P)\vee (S\wedge \neg T\wedge P\wedge I)$.\\
$\triangle_{\mu}^{GMax}=\neg S\wedge \neg I\wedge ((\neg T\wedge
P)\vee (T\wedge \neg P))$.

However, the results are not intuitive from the view of
propositional symbols. For variable $I$, since $\varphi_1,\varphi_2$
don't care about it and $\varphi_3,\varphi_4$ support $\neg I$, So
the literal formula $\neg I$ can be regarded as the perspective of
the whole group. For variables $S,T,P$, the former two owners and
the third one have different opinions. That is, the former two
support $S,T$ and $P$ and the third one supports their opposite. So
they don't reach consensus if no co-owner gives in. The formula
$\neg I\wedge \mu$ is thus natural result for the merging. Moreover,
$\triangle_{\mu}^{\Sigma}(\Phi)$ and $\triangle_{\mu}^{Max}$ are not
enough to capture $\neg I$. on the other hand, although
$\triangle_{\mu}^{GMax}\models \neg I$, the operator is too strong
since $\triangle_{\mu}^{GMax}\models \neg S$. Two co-owners support
$S$, one supports its opposite and one is neutral about $S$. We
can't agree on $\neg S$ even if we don't follow the majority
property. It is reasonable that $S$ is taken as unknown, i.e.,
neither $S$ nor $\neg S$.

In general, inconsistencies occurring between KBs are caused by
over-defining in representing something. The main idea of handling
inconsistency is removing or ignoring those redundant information.
As a significant approach to dealing with inconsistencies,
forgetting is a useful tool to restrict variables to be discussed in
a subset of variables of the original KBs with keeping logical
equivalence locally (\cite{Lin94forgetit!}).

In this paper, inspiring from \cite{lang2010reasoning}, we present a
novel approach based on variable forgetting to merging multi-KBs
with maintaining consistency. First, we build the relationship
between belief merging and variable forgetting by using the
framework of dilation (presented in \cite{gorogiannis2008merging}).
We then reformalize three classical merging operators via variable
forgetting. Two new forgetting-based merging operators are obtained
by modifying variables to be forgotten. We show that those
forgetting-based merging operators can amend the shortage (discussed
in the motivating example of ``swimming-pool'') of classical
operators.

The rest of this paper is organized as follows. Section 2 gives a
brief review of merging and forgetting for KBs. Section 3 employs
forgetting to capture traditional merging operators. We develop
three new merging operators by forgetting variables in Section 4. in
the last section, we summarize this paper and put forward to some future
works.

\section{Preliminaries}
In this section, we briefly review basic concepts of belief merging
and forgetting for propositional KBs.
\subsection{Belief Merging}
The propositional language, denoted by $\mathcal{L}$, is constructed
from a finite set $\mathcal{P}$ of symbols. $\top$(true) and
$\bot$(false) are boolean constants. In $\mathcal{L}$, $p$ denotes a
propositional variable, $\varphi,\psi,\mu$ propositional formula,
$\Phi,\Phi_1,\Phi_2,\cdots$ sets of formulas. $\Phi_1\cup \Phi_2$ is
the union of sets $\Phi_1, \Phi_2$, and $\Phi_1\sqcup \Phi_2$ is the
union of multi-sets of sets $\Phi_1, \Phi_2$. $Var(\varphi)$ denotes
the set of variables which occur in $\varphi$. An interpretation is
a function from $\mathcal{P}$ to $\{0,1\}$. $\omega,\omega'$ denote
interpretations. $\mathcal{M}$ is the collection of all
interpretations. An interpretation is a model of formula iff it
makes it true in classical way. $mod(\varphi)$ denotes the set of
models of $\varphi$. $\varphi$ is consistent iff $mod(\varphi)\neq
\emptyset$. $\varphi\equiv\psi$ iff $mod(\varphi)=mod(\psi)$.

The aggregation of finite KBs into a collective one is studied by a
recent discipline called \emph{belief merging} (see
\cite{silberschatz1990database,konieczny2002merging,gorogiannis2008merging}).
A particular type of aggregation is called \emph{model-based
merging}. Intuitively, the model-based merging is aggregating those
models which are closer to models of every formula. In technique, it
is choosing those models which have the minimal aggregating
distance.

Let $\Phi'=\{\varphi_1',\cdots,\varphi_n'\}$, $\Phi\leftrightarrow
\Phi'$ iff there is a bijection $f$: $\Phi\rightarrow\Phi'$
satisfying $\forall \varphi_i\in \Phi$, $\varphi_i\leftrightarrow
f(\varphi_i)$.

$|V|$ denotes the cardinal number of a set $V$. The (Dalal) distance
between two models $d(\omega,\omega')=\mid\{p\mid\omega,\omega'$
assign differently on $p\}\mid$. The distance between a model and
formula is defined as $d(\omega,\varphi)=min_{\omega'\in
mod(\varphi)}d(\omega,\omega')$.

$\varphi_{x\leftarrow 0}$(resp. $\varphi_{x\leftarrow 1}$) denotes
the formula obtained by replacing in $\varphi$ every occurrence of
variables $x$ by $\bot$(resp. $\top$).

Let $\Phi=\{\varphi_1,\cdots,\varphi_n\}$, $\wedge
\Phi=\wedge_{\varphi_i\in \Phi}\varphi_i$. Formulas
$\varphi,\varphi'$ denote KBs, $\mu,\mu_1,\mu_2$ represent the
integrity constraints. For simple discussion, we continue to assume
every formula $\varphi_i\in \Phi$ is consistent in this paper.

There are nine postulates (\textbf{IC0})-(\textbf{IC8}) presented in
\cite{konieczny2002merging} to capture the belief merging.

Let $\triangle$ be an $IC$ merging operator iff it
satisfies the following postulates:\\
(\textbf{IC0}) $\triangle_{\mu}(\Phi)\vdash \mu$\\
(\textbf{IC1}) If $\mu$ is consistent, then $\triangle_{\mu}(\Phi)$ is
consistent\\
(\textbf{IC2}) If $\wedge\Phi$ is consistent with $\mu$, then
$\triangle_{\mu}(\Phi)=\wedge\Phi\wedge \mu$\\
(\textbf{IC3}) If $\Phi_1\leftrightarrow\Phi_2$ and
$\mu_1\leftrightarrow\mu_2$, then
$\triangle_{\mu_1}(\Phi_1)\leftrightarrow
\triangle_{\mu_2}(\Phi_2)$\\
(\textbf{IC4}) If $\varphi\vdash\mu$ and $\varphi'\vdash\mu$, then
$\triangle_{\mu}(\varphi\sqcup\varphi')\wedge\varphi\not\vdash
\bot\Rightarrow
\triangle_{\mu}(\varphi\sqcup\varphi')\wedge\varphi'\not\vdash
\bot$\\
(\textbf{IC5}) $\triangle_{\mu}(\Phi_1)\wedge \triangle_{\mu}(\Phi_2)\vdash
\triangle_{\mu}(\Phi_1\sqcup \Phi_2)$\\
(\textbf{IC6}) If $\triangle_{\mu}^{\Phi_1}\wedge \triangle_{\mu}(\Phi_2)$ is
consistent, then $\triangle_{\mu}(\Phi_1\sqcup \Phi_2)\vdash
\triangle_{\mu}(\Phi_1)\wedge \triangle_{\mu}(\Phi_2)$\\
(\textbf{IC7}) $\triangle_{\mu_1}(\Phi)\wedge \mu_2\vdash
\triangle_{\mu_1\wedge\mu_2}(\Phi)$\\
(\textbf{IC8}) if $\triangle_{\mu_1}(\Phi)\wedge \mu_2$ is consistent, then
$\triangle_{\mu_1\wedge\mu_2}(\Phi)\vdash
\triangle_{\mu_1}(\Phi)\wedge \mu_2$

Besides (\textbf{IC0})-(\textbf{IC8}), there are some additional postulates to
characterize the other properties as follows:

%
%

The \emph{majority property} (\textbf{Maj}):  $\exists n \triangle_{\mu}
(\Phi_1\sqcup\Phi_2^n)\vdash \triangle_{\mu} (\Phi_2)$. Intuitively,
if a subgroup appears enough in the whole group, then it is the
opinion of the group.

\emph{Majority independence} requires us only consider different
KBs. It is denoted by (\textbf{MI}): $\triangle_{\mu}
(\Phi_1\sqcup\Phi_2^n)\leftrightarrow \triangle_{\mu}
(\Phi_1\sqcup\Phi_2)$.


There are three traditional model-based merging operators as
follows: $\triangle_{\mu}^{\Sigma}$ (presented in
\cite{lin1999knowledge}), $\triangle_{\mu}^{Max}$ (presented  in
\cite{revesz1993semantics}) and $\triangle_{\mu}^{GMax}$ (presented
in \cite{konieczny2002merging}).

Let $\Phi$ be a KB and $\omega,\omega'$ two interpretations.
\begin{itemize}
  \item The $\Sigma-$distance between an interpretation and a KB
is defined as $d_{\Sigma}(\omega,\Phi)=\Sigma_{\varphi\in
\Phi}d(\omega,\varphi)$. Then we have the following pre-order:
$\omega\leq_{\Phi}^{\Sigma}\omega'$ iff $d_{\Sigma}(\omega,\Phi)\leq
d_{\Sigma}(\omega',\Phi)$. The merging operator
$\triangle_{\mu}^{\Sigma}$ is defined:
$mod(\triangle_{\mu}^{\Sigma})=min(mod(\mu),\leq_{\Phi}^{\Sigma})$.

It easily shows that $\triangle_{\mu}^{\Sigma}$ satisfies
(\textbf{IC0})-(\textbf{IC8}), (\textbf{Maj}).
\item The $Max-$distance between an interpretation and a KB
is defined as follows: $d_{Max}(\omega,\Phi)=Max_{\varphi\in
\Phi}d(\omega,\varphi)$. Then we have the following pre-order:
$\omega\leq_{\Phi}^{Max}{\omega'}$ iff $d_{Max}(\omega,\Phi)\leq
d_{Max}(\omega',\Phi)$. The merging operator $\triangle_{\mu}^{Max}$
is defined as follows:
$mod(\triangle_{\mu}^{Max})=min(mod(\mu),\leq_{\Phi}^{Max})$.

It easily shows that $\triangle_{\mu}^{Max}$ satisfies (\textbf{IC0})-(\textbf{IC5}),
(\textbf{IC7}), (\textbf{IC8}) and (\textbf{MI}). In particular, it can't satisfy (\textbf{IC6}), (\textbf{Maj}).

\item Suppose $\Phi=\{\varphi_1,\cdots,\varphi_n\}$.
$d_{j}^{\omega}=d(\omega,\varphi_j)$. Let $L_{\omega}^{\Phi}$ be the
list obtained from $(d_1^\omega,\cdots,d_n^\omega)$ by sorting it in
descending order. Let $\leq_{lex}$ be the lexicographical order
between sequences of integers. Then the pre-order
$\leq_{\Phi}^{GMax}$ is defined as follows:
$\omega\leq_{\Phi}^{GMax}\omega'$ iff
$L_\omega^{\Phi}\leq_{lex}L_{\omega'}^{\Phi}$. The merging operator
$\triangle_{\mu}^{GMax}$ is defined as follows:
$mod(\triangle_{\mu}^{GMax}=min(mod(\mu),\leq_{\Phi}^{GMax})$.

It easily shows that $\triangle_{\mu}^{GMax}$ satisfies (\textbf{IC0})-(\textbf{IC8}).
\end{itemize}

Some other merging operators such as DA$^2$ presented in
\cite{konieczny2004} could be taken as extensions of three classical
operators.

\subsection{Forgetting}
Forgetting proposed by Lin and Reiter (\cite{Lin94forgetit!}) is
filtering all facts that are no longer true from KBs. That is to
say, forgetting is taken as a basic operation for weakening
formulas. In this sense, variables forgetting could be employed to
reason under inconsistency (\cite{ lang2010reasoning}).

Let $\varphi$ be a propositional formula, $p$ be an atom and $V$ be
a set of variables. $\exists V.\varphi$ denotes forgetting $V$ in
$\varphi$ which is recursively defined as follows:
\begin{itemize}
\item $\exists\emptyset.\varphi\equiv \varphi$;
\item $\exists\{p\}.\varphi\equiv \varphi_{p\leftarrow 0}\vee \varphi_{p\leftarrow
1}$;
\item $\exists (V \cup \{p\}).\varphi\equiv \exists V.(\exists
\{p\}.\varphi)$.
\end{itemize}

Let $switch(\omega,p)$ denote the interpretation that assigns the
same truth values to all variables
except $p$, and assigns the opposite value to $p$. Then\\
$mod(\exists\{p\}.\varphi)=mod(\varphi)\cup\{switch(\omega,p)\mid\omega\models
\varphi\}$

Next we enumerate some good properties of forgetting for knowledge
bases which will be useful for our work.
\begin{itemize}
\item Let $\varphi$ be consistent, $V_1\subseteq V_2\subseteq
Var(\varphi)$, then $\exists V_1.\varphi\models \exists
V_2.\varphi$. In particular $\varphi\models \exists V.\varphi$.
\item $V=Var(\varphi)$. If $\varphi$ is consistent, then
$\exists V.\varphi\models\top$; If $\varphi$ isn't consistent, then
$\exists V.\varphi\models\bot$.
\item Let $\varphi$ be consistent, if $Var(\varphi)\subseteq V$, then
$\exists V.\varphi\models\top$.
\item Let $\varphi,\psi$ be two formulas, $V$ be a variables set, then
$\exists V.(\varphi\vee\psi)\equiv\exists V.\varphi\vee \exists
V.\psi$.
\item If $p\not\in Var(\varphi)$, then $\exists \{p\}.\varphi\equiv
\varphi$.
\item If $\varphi\models \varphi'$ then $\exists V.\varphi\models \exists
V.\varphi'$. In particular, if $Var(\varphi')\cap V=\emptyset$ then
$\exists V.\varphi\models \varphi'$.
\end{itemize}

\section{Relationship Between Merging and Forgetting}
In this section, we apply variable forgetting to capture three
existing merging operators
$\triangle_{\mu}^{\Sigma},\triangle_{\mu}^{Max},\triangle_{\mu}^{GMax}$.
At first, we need introduce a so-called \emph{operator of dilation}
(presented in \cite{gorogiannis2008merging}) to build the inner
relationship between belief merging and variable forgetting.

A dilation operator $D$ is a mapping from formula to formula
satisfying: $mod(D(\varphi))=\{\omega\in \mathcal{M}\mid
d(\omega,\varphi)\leq 1\}$.

We have $D^1(\varphi)=D(\varphi)$ and
$D^n(\varphi)=D(D^{n-1}(\varphi))$. So
$mod(D^n(\varphi))=\{\omega\in \mathcal{M}\mid d(\omega,\varphi)\leq
n\}$.

Let $\Phi=\{\varphi_1,\varphi_2,\cdots,\varphi_n\}$. These operators
can be equivalently expressed by dilation.
\begin{itemize}
\item $\triangle_{\mu}^{\Sigma}(\Phi)\equiv\vee_{c_1+\cdots+c_n=k}(D^{c_1}(\varphi_1)\wedge \cdots\wedge D^{c_n}(\varphi_n)\wedge \mu)$, where $k$ is the least number so that
the disjunction is consistent.
\item
$\triangle_{\mu}^{Max}(\Phi)\equiv D^{k}(\varphi_1)\wedge
\cdots\wedge D^{k}(\varphi_n)\wedge\mu$, where $k$ is the least
number so that $D^{k}(\varphi_1)\wedge \cdots\wedge
D^{k}(\varphi_n)\wedge\mu$ is consistent.
\item $\triangle_{\mu}^{GMax}(\Phi)\equiv\vee_{<c_1,\cdots,c_n>\in perm(T)}(D^{c_1}(\varphi_1)\wedge \cdots\wedge D^{c_n}(\varphi_n)\wedge
\mu)$, where $T$ is an n-tuple of integers, which is sorted in
descending order, is lexicographically least such that the
disjunction is consistent.
\end{itemize}

Next we present that dilation can be captured by forgetting. Before
presenting, a key lemma will be stated in the following.

\begin{lem}
Let $\varphi$ be a formula and $V$ be a set of variables. We have
$\forall \omega\in mod(\exists V.\varphi)$, $d(\omega,\varphi)\leq
|V|$.
\end{lem}
\begin{IEEEproof} $\forall \omega\in mod(\exists V.\varphi)$, there must be
$\omega'\in mod(\varphi)$ which satisfies $\omega,\omega'$ assume
the same on variables except for variables of $V$. So
$d(\omega,\omega')\leq |V|$. Thus $d(\omega,\varphi)\leq |V|$.
 \end{IEEEproof}

\begin{thm}
Let $\varphi$ be a consistent formula, $p$ be a propositional
variable, $V$ be a set of variables and $n$ be an integer. We have
\begin{itemize}
  \item $D(\varphi)\equiv\vee_{p\in Var(\varphi)}\exists \{p\}.\varphi$.
  \item If $1<n\leq |Var(\varphi)|$, then
$D^n(\varphi)\equiv\vee_{V\subseteq Var(\varphi),|V|=n}\exists
V.\varphi$.
  \item If $n>|Var(\varphi)|$, then $D^n(\varphi)\equiv\top$.
\end{itemize}
\end{thm}

\begin{IEEEproof} 1. We will prove they have the same models. $\forall \omega\in
mod(D(\varphi))$, then $d(\omega,\varphi)\leq 1$. If
$d(\omega,\varphi)=0$ then $\omega\in mod(\varphi)$. So $\omega\in
mod(\exists \{p\}.\varphi)$ and $\omega\in mod(\vee_{p\in
Var(\varphi)}\exists \{p\}.\varphi)$. If $d(\omega,\varphi)=1$, then
$\exists \omega'\in mod(\varphi)$ and $d(\omega,\omega')=1$. So
$\omega,\omega'$ interpret the same except $p$ in $var(\varphi)$.
Thus $\omega\in mod(\exists \{p\}.\varphi)$. Otherwise, if
$\omega\in mod(\vee_{p\in Var(\varphi)}\exists \{p\}.\varphi)$ then
$\omega\in mod(\exists \{p\}.\varphi)$ for some $p$. So
$d(\omega,\varphi)\leq 1$ holds by the lemma.

2. If $1<n\leq |Var(\varphi)|$, Then
$D^n(\varphi)=D(D^{n-1}(\varphi))$. Suppose
$D^{n-1}(\varphi)=\vee_{V\subseteq Var(\varphi),|V|=n-1}\exists
V.\varphi$. And $D^n(\varphi)=D(\vee_{V\subseteq
Var(\varphi),|V|=n-1}\exists V.\varphi)\\
 =\vee_{p\in
Var(\varphi)}\exists\{p\}.(\vee_{V\subseteq
Var(\varphi),|V|=n-1}\exists V.\varphi)\\
 =\vee_{V\subseteq
Var(\varphi),|V|=n}\exists V.\varphi\vee \vee_{V\subseteq
Var(\varphi),|V|=n-1}\exists V.\varphi\\
=\vee_{V\subseteq Var(\varphi),|V|=n}\exists V.\varphi$.

The last equation is because of a property of forgetting. Actually,
$D^n(\varphi)=\vee_{V\subseteq Var(\varphi),|V|\leq
n}forget(\varphi,V)$. And we can omit $V\in Var(\varphi)$ in
subscript without confusion, i.e., $D^n(\varphi)=\vee_{|V|\leq
n}forget(\varphi,V)$.

3. If $n>|Var(\varphi)|$, then $mod(D^n(\varphi))=\{\omega\in
\mathcal{M}|d(\omega,\varphi)\leq n\}$. $\exists
Var(\varphi).\varphi\equiv \top$, then $\forall \omega\in
\mathcal{M}$, then $\omega\in mod(\exists Var(\varphi).\varphi)$.
Thus $d(\omega,\varphi)\leq |Var(\varphi)|<n$. So $\omega \in
mod(D^n(\varphi))$ and $D^n(\varphi)=\top$.
\end{IEEEproof}

Since the relationship between dilation and forgetting is pointed
out, it's natural to represent these operators by forgetting.

\begin{thm}
Let $\Phi=\{\varphi_1,\varphi_2,\cdots,\varphi_n\}$.
\begin{itemize}
  \item $\triangle_{\mu}^{\Sigma}(\Phi)\equiv\vee_{|V_1|+\cdots+|V_n|=k}(\exists
V_1.\varphi_1\wedge\cdots\wedge\exists V_n.\varphi_n \wedge\mu)$.
  \item $\triangle_{\mu}^{Max}(\Phi)\equiv\vee_{|V_1|=\cdots=|V_n|=k}(\exists
V_1.\varphi_1\wedge\cdots\wedge\exists V_n.\varphi_n \wedge\mu)$.
  \item $\triangle_{\mu}^{GMax}(\Phi)=\vee_{<|V_1|,\cdots,|V_n|>\in
perm(T)}(\exists V_1.\varphi_1\wedge\cdots\wedge\exists
V_n.\varphi_n \wedge\mu)$.
\end{itemize}
Here $k$ and $T$ are the same as above definitions.
\end{thm}

\begin{IEEEproof}
We only prove the first equality and the others are similar. Let
$\triangle_{\mu}^{\Sigma}(\Phi)\equiv
\vee_{c_1+\cdots+c_n=k}(D^{c_1}(\varphi_1)\wedge \cdots\wedge
D^{c_n}(\varphi_n)\wedge \mu)$ and $k$ is the least number that it
is consistent. Then $\triangle_{\mu}^{\Sigma}(\Phi)\equiv
\vee_{c_1+\cdots+c_n=k}(\vee_{|V_1|=c_1}(\exists
V_1.\varphi_1)\wedge \cdots \wedge \vee_{|V_n|=c_n}(\exists
V_n.\varphi_n)\wedge \mu)\equiv
\vee_{c_1+\cdots+c_n=k}(\vee_{|V_1|=c_1,\cdots,|V_n|=c_n}(\exists
V_1.\varphi_1\wedge\cdots\wedge \exists V_n.\varphi_n\wedge
\mu)\equiv \vee_{|V_1|+\cdots+|V_n|=k}(\exists
V_1.\varphi_1\wedge\cdots\wedge\exists V_n.\varphi_n \wedge\mu)$.
\end{IEEEproof}

A major difference between dilation and forgetting is that every
variable has to be forgotten in dilation as it only find all the
closest models to the original formula.

\section{Revised Merging Operators Using Forgetting}
The merging operators are represented by forgetting have the similar
forms as by dilation. Since they have so close relationship that we
can construct merging operators directly by forgetting. Our method
is to revise the merging operators represented by forgetting to
obtain some new ones which satisfy two new properties.

Theorem 2 shows that forgotten variables for different knowledge
bases might be distinct. Thus the result of conjunction after
forgetting doesn't focus on a special domain. If we restrict every
KB to forget the same set of variables, we get a new operator
defined in this domain. Of course, we consider the minimal sets of
variables to make the result of KBs after forgetting consistent.

Literal formulas might be regarded as the simplest sub-language
which is constructed on a single atom. In the face of literal
formula query, all other variables needn't to be considered. As the
motivating example presented in Section 1, on the one hand, if some
information sources entail a literal formula, but the others say
nothing about this atom, then the literal formula should hold for
the group because they don't conflict with each other about this
atom. On the other hand, if some sources agree on a literal formula,
but some sources object to it, then it should be rejected. So our
attitude to merging is more skeptical than before based on this
point.

Next we formalize such two properties.

(\textbf{A1}) Let a KB $\Phi'\subseteq \Phi$ and $l$ be a literal
formula. Suppose $\mu\wedge l$ is consistent. If $\forall
\varphi'\in \Phi', \varphi'\models l$ and $Var(l)\not\in
Var(\Phi-\Phi')$, then the merging result
$\triangle_{\mu}(\Phi)\models l\wedge \mu$.

(\textbf{A2}) Let $\varphi_1,\varphi_2\in\Phi$. If $\varphi_1\models
l$ and $\varphi_2\models \neg l$, then $\triangle_{\mu}(\Phi)\not
\models l$. The property (\textbf{A2}) implicitly requires that
$\triangle_{\mu}(\Phi)\not \models \neg l$ holds too.

Though a KB $\Phi$ is inconsistent w.r.t $Var(\Phi)$, it may be
consistent w.r.t some subset of $Var(\Phi)$. We consider all these
maximal subsets on set-inclusion or cardinal number and get two new
operators in the following.

\begin{defn}
Let $\Phi=\{\varphi_1,\varphi_2,\cdots,\varphi_n\}$ and $\mu$ be an
integrity constraint. $V\subseteq Var(\Phi)$, the collection of
minimal variables sets for forgetting $FS=\{V$ is minimal w.r.t set
cardinal $\mid \wedge_{i=1}^{n}\exists V.\varphi_i\wedge \mu$
consistent\}. We define $\triangle_{\mu}^{f_1}$ as follows:
$\triangle_{\mu}^{f_1}(\Phi)= \vee_{V\in FS}(\wedge_{i=1}^{n}\exists
V.\varphi_i\wedge \mu).$
\end{defn}

The new operator satisfies (\textbf{A1}) and (\textbf{A2}). Before we prove the two
properties above, we need the following lemma.
\begin{lem}
Let $p$ be variable in $V$, $l$ be a literal formula on $p$. If
$\forall 1\leq i\leq n, \varphi_i\models l$, and $\wedge_{i=1}^n
\exists V.\varphi_i$ is consistent, then $\wedge_{i=1}^n \exists
(V-\{p\}).\varphi_i$ is also consistent.
\end{lem}

\begin{IEEEproof}
$\forall 1\leq i\leq n, \varphi_i\models l$, so we obtain $\exists
(V-\{p\}).\varphi_i\models l$. Let $\omega\in mod(\wedge_{i=1}^n
\exists V.\varphi_i)$, then $\omega\in mod(\exists V.\varphi_i)$. If
$\omega\models l$ then $\omega\in mod(\exists (V-\{p\}).\varphi_i)$,
and $\omega\in mod(\wedge_{i=1}^n \exists (V-\{p\}).\varphi_i)$. So
it is consistent. If $\omega\models \neg l$, then there is a model
$\omega'\in mod(\wedge_{i=1}^n \exists V.\varphi_i)$ satisfying
$\omega'$ and $\omega$ give the same truth value except $p$. It can
be reduced to the first case and $\omega'\in mod(\wedge_{i=1}^n
\exists (V-\{p\}).\varphi_i)$. Thus $\wedge_{i=1}^n \exists
(V-\{p\}).\varphi_i$ is also consistent.
\end{IEEEproof}

The lemma indicates that the minimal forgetting set of variables
doesn't comprise the variables on which all KBs agree.
It can be easily extended to the general case that $p$ doesn't occur
in some bases and the other bases including it entail $l$.
\begin{thm}
$\triangle_{\mu}^{f_1}$ satisfies properties (\textbf{A1}), (\textbf{A2}).
\end{thm}
\begin{IEEEproof}
(\textbf{A1}) Let $V'=Var(\Phi)-\{p\}$. If $\forall \varphi'\in \Phi',
\varphi'\models p$, then $\forall \varphi'\in \Phi', \exists
V'.\varphi'\models p$. So $\exists V'.\varphi'\equiv p$. And
$\forall \varphi\in \Phi-\Phi', \exists V'.\varphi\equiv \top$. Thus
if $\mu\not \models \neg p$ then $\wedge_{\varphi\in \Phi}(\exists
V'.\varphi)\wedge\mu \equiv p \wedge \mu$ is consistent. In the
definition of $\triangle_{\mu}^{f_1}$, $\forall V\in FS, p\not\in
V$. Otherwise, $V-\{p\}$ is minimal variables set such that
$\wedge_{\varphi\in \Phi}(\exists (V-\{p\}).\varphi)\wedge\mu$ is
consistent. Thus $\triangle_{\mu}^{f_1}(\Phi)\models p$.

(\textbf{A2}) $\forall V\in FS, p\in V$ holds. Otherwise, suppose $p\not \in
V$. Let $\varphi_1\models p$ and $\varphi_2\models \neg p$. So
$\exists V.\varphi_1\models p$ and $\exists V.\varphi_2\models \neg
p$. $\exists V.\varphi_1\wedge\exists V.\varphi_2$ is inconsistent.
$V\not\in FS$. The property is proved.
\end{IEEEproof}

We return to consider the motivating example. $\varphi_1(\varphi_2)$
and $\varphi_3$ assign contrary truth values on variables $S,T,P$.
So they should forget these variables to preserve consistency.
$V=\{S,T,P\}$ is the only element in $FS$ of definition above,
$\triangle_{\mu}^{f_1}(\Phi)=\wedge_{i=1}^4 \exists
V.\varphi_i\wedge \mu=\neg I\wedge\mu=\neg I\wedge(\neg S\vee \neg
T)\wedge (\neg S\vee \neg P)\wedge (\neg P\vee \neg T)$.

Before we prepare to show properties of $\triangle_{\mu}^{f_1}$, we
first introduce an important lemma.

\begin{lem}
Let $\varphi,\varphi'$ be two formulas and $V$ a set of variables.
We have $d(\exists V.\varphi,\varphi')=d(\exists V.\varphi,\exists
V.\varphi')$.
\end{lem}

\begin{IEEEproof}
We prove first $d(\exists \{p\}.\varphi,\varphi')=d(\exists
\{p\}.\varphi,\exists \{p\}.\varphi')$ for any atom $p$.
$Mod(\varphi')\subseteq Mod(\exists \{p\}.\varphi')$, so $d(\exists
\{p\}.\varphi,\exists \{p\}.\varphi')\leq d(\exists
\{p\}.\varphi,\varphi')$. Take $\omega\in Mod(\exists
\{p\}.\varphi),\omega'\in Mod(\exists \{p\}.\varphi')$,
$d(\omega,\omega')=d(\exists \{p\}.\varphi,\exists \{p\}.\varphi')$.
There are four
cases: \\
1. If $\omega\in Mod(\varphi),\omega'\in Mod(\varphi')$, then the
equation holds.\\
2. If $\omega\in Mod(\varphi),\omega'\in Mod(\exists
\{p\}.\varphi')-Mod(\varphi')$, let $\omega_1,\omega_1'$ assign
differently only on $p$ with $\omega,\omega'$ respectively. Then
$d(\omega_1,\omega_1')=d(\omega,\omega')$ and $\omega_1\in
Mod(\exists \{p\}.\varphi),\omega_1'\in
Mod(\varphi')$. So the equation holds.\\
3. If $\omega\in Mod(\exists \{p\}.\varphi)-Mod(\varphi),\omega\in
Mod(\varphi')$, then the equation holds.\\
4. If $\omega\in Mod(\exists \{p\}.\varphi)-Mod(\varphi),\omega'\in
Mod(\exists \{p\}.\varphi')-Mod(\varphi')$, Let $\omega_1,\omega_1'$
assign differently only on $p$ with $\omega,\omega'$ respectively.
Then $d(\omega_1,\omega_1')=d(\omega,\omega')$ and $\omega_1\in
Mod(\varphi),\omega_1'\in Mod(\varphi')$. So the equation holds.

Next we prove the general case for any set $V(|V|>1)$. Let
$V=\{p_1,\cdots,p_n\}$. $d(\exists V.\varphi,\exists V.\varphi')\\
=d(\exists \{p_1\}.(\exists (V-\{p_1\}).\varphi),\exists \{p_1\}.(\exists (V-\{p_1\}).\varphi'))\\
=d(\exists \{p_1\}.(\exists (V-\{p_1\}).\varphi),\exists (V-\{p_1\}).\varphi')\\
=d(\exists V.\varphi,\exists (V-\{p_1\}).\varphi')\\
=d(\exists V.\varphi,\exists (V-\{p_1,p_2\}).\varphi')\\
= \cdots\\
=d(\exists V.\varphi,\varphi')$.
\end{IEEEproof}

This lemma shows $d(\exists V.\varphi,\varphi')=d(\exists
V.\varphi',\varphi)$. So $\exists V.\varphi$ is consistent with
$\varphi'$ iff $\exists V.\varphi'$ is consistent with $\varphi$.

It's easy to see that for any $V'\subseteq V$, $d(\exists
V.\varphi,\varphi')\equiv d(\exists V.\varphi,\exists V'.\varphi')$.
But $\exists V.\varphi\wedge \varphi'\equiv\exists V.\varphi\wedge
\exists V'.\varphi'$ doesn't hold. For example, $\varphi\equiv
p\wedge q,\varphi'\equiv\neg p$. $\exists \{p\}.\varphi\wedge
\varphi'\equiv q\wedge \neg p$, but $\exists \{p\}.\varphi\wedge
\exists \{p\}.\varphi'\equiv q$.

\begin{thm} $\triangle_{\mu}^{f_1}(\Phi)$ satisfies
postulates (\textbf{IC0})-(\textbf{IC4}), (\textbf{IC7}), (\textbf{IC8}) and (\textbf{MI}).
\end{thm}

\begin{IEEEproof} 1. (\textbf{IC0})-(\textbf{IC3}) obviously hold for
$\triangle_{\mu}^{f_1}(\Phi)$.

2. $\triangle_{\mu}^{f_1}(\varphi\cup \varphi')\wedge
\varphi=\vee_{V\in FS}(\exists V.\varphi\wedge \exists
V.\varphi'\wedge \mu) \wedge \varphi= \vee_{V\in FS}(\varphi\wedge
\exists V.\varphi')$ and $\triangle_{\mu}^{f_1}(\varphi\cup
\varphi')\wedge \varphi'=\vee_{V\in FS}(\varphi'\wedge \exists
V.\varphi)$. According to lemma 3 above, $\exists V.\varphi\wedge
\exists V.\varphi'$ is consistent if and only if $\exists
V.\varphi\wedge \varphi'$ (or $\exists V.\varphi'\wedge \varphi$) is
consistent. So (\textbf{IC4}) holds.

3. Now we prove (\textbf{IC7}) and (\textbf{IC8}). $FS,FS'$ are collections of
forgetting set for $\triangle_{\mu_1}(\Phi)$ and
$\triangle_{\mu_1\wedge\mu_2}(\Phi)$ respectively in the definition.
Let $\triangle_{\mu_1}^{f_1}(\Phi)=\vee_{V\in
FS}(\wedge_{i=1}^{n}\exists V.\varphi_i\wedge\mu_1)$. If
$\triangle_{\mu_1}^{f_1}(\Phi)\wedge \mu_2$ is consistent, then
$\exists V_0$ satisfying that $\wedge_{i=1}^{n}\exists
V_0.\varphi_i\wedge\mu_1\wedge\mu_2$ is consistent. The cardinals of
sets in $FS,FS'$ are the same. So $V_0\in FS'$. Conversely, $\forall
V'\in FS'$, $V'\in FS$. Then we have
$\triangle_{\mu_1}^{f_1}(\Phi)\wedge\mu_2 \equiv
\triangle_{\mu_1\wedge\mu_2}^{f_1}(\Phi)$.

4. For some set $V$, the conjunction of every formula in
$\Phi_1\sqcup\Phi_2^n$ after forgetting $V$ is consistent if and
only if it is the same case for $\Phi_1\sqcup\Phi_2$ after
forgetting $V$. The collection of forgetting sets in
$\triangle_{\mu}^{f_1}(\Phi_1\sqcup\Phi_2^n)$is the same as the one
for $\triangle_{\mu}^{f_1}(\Phi_1\sqcup\Phi_2)$.
The knowledge set $\Phi_1\sqcup \Phi_2^n$ after forgetting $V\in FS$ is\\
$\vee_{V\in FS}(\wedge_{\varphi\in
\Phi_1}\exists V.\varphi\wedge_{i=1}^n(\wedge_{\psi\in\Phi_2}\exists V.\psi)\wedge\mu)\\
=\vee_{V\in FS}(\wedge_{\varphi\in \Phi_1}\exists
V.\varphi\wedge_{\psi\in\Phi_2}\exists V.\psi\wedge\mu)$. Thus after
forgetting $V$, two knowledge sets $\Phi_1\sqcup \Phi_2^n$ and
$\Phi_1\sqcup \Phi_2$ become identical, i.e.,
$\triangle_{\mu}^{f_1}(\Phi_1\sqcup
\Phi_2^n)\equiv\triangle_{\mu}^{f_1}(\Phi_1\sqcup \Phi_2)$.
\end{IEEEproof}

The operator $\triangle_{\mu}^{f_1}$ doesn't satisfy postulates
(\textbf{IC5}) and (\textbf{IC6}), but it satisfies the proposed
properties (\textbf{A1}) and (\textbf{A2}). We think that
$\triangle_{\mu}(\Phi_1)\wedge \triangle_{\mu}(\Phi_2)$ doesn't
represent the common alternatives of the two groups which are indeed
$Th(\triangle_{\mu}(\Phi_1))\cap Th(\triangle_{\mu}(\Phi_2))$ (
$Th(\varphi)$ denotes the deduction closure of formula $\varphi$).
If we select the minimal sets w.r.t set-inclusion then we get
another operator.

\begin{defn}
Let $\Phi=\{\varphi_1,\varphi_2,\cdots,\varphi_n\}$ and $\mu$ an
integrity constraint. $V\subseteq \cup_{i=1}^{n}Var(\varphi_i)$, the
minimal number of variables for forgetting $FS=\{V$ is minimal w.r.t
set-inclusion $\mid \wedge_{i=1}^{n}\exists V.\varphi\wedge \mu$ is
consistent\}. We define an operator as follows:
$\triangle_{\mu}^{f_2}(\Phi)= \vee_{V\in FS}(\wedge_{i=1}^{n}\exists
V.\varphi\wedge \mu).$
\end{defn}

For the example of co-owners, The results of $\triangle_{\mu}^{f_2}$
and $\triangle_{\mu}^{f_1}$ are the same one. Analogously,
$\triangle_{\mu}^{f_2}$ satisfies (\textbf{A1}), (\textbf{A2}) and some postulates.

\begin{thm} $\triangle_{\mu}^{f_2}$ satisfies
 (\textbf{A1}), (\textbf{A2}), (\textbf{IC0})-(\textbf{IC4}), (\textbf{IC7}) and (\textbf{MI}).
\end{thm}

The next theorem states that the property (\textbf{IC8}) doesn't
hold, thereby $\triangle_{\mu}^{f_1}$ and $\triangle_{\mu}^{f_2}$
are different. For example, Let $\Phi=\{\varphi_1,\varphi_2\}$,
$\mu=\top$ and $\varphi_1=\neg p\wedge \neg q\wedge \neg r\wedge\neg
s$, $\varphi_2=((p\wedge\neg q\wedge\neg r)\vee (\neg p\wedge
q\wedge r))\wedge \neg s$. $\{a\},\{b,c\}$ are the forgetting sets
for $\triangle_{\mu}^{f_2}$, but $\{a\}$ is the only one for
$\triangle_{\mu}^{f_1}$. So $\triangle_{\mu}^{f_1}(\Phi)\equiv \neg
q\wedge \neg r\wedge \neg s$, $\triangle_{\mu}^{f_2}\equiv \neg
p\wedge \neg s$.

These two operators belong to homogeneous context in
\cite{lang2010reasoning} in which they propose three contexts for
forgetting.

\section{Conclusion and Future Work}
This paper proposes a scenario to eliminate conflicts occurring in
the process of merging KBs by applying variable forgetting. Firstly,
we discuss the relationship between belief merging and variable
forgetting for KBs via the operation of dilation. As an interesting
result, three classical model-based merging operators can be well
captured by variable forgetting. Based on this relationship, we
revise those merging operators by modifying variables in forgetting
so that these new operators (after revising) become more smart in
managing multiple KBs. Though our work is inspired from
\cite{lang2010reasoning}, our operator is based on variable
selection on multiple KBs while \cite{lang2010reasoning} is based on
the context of singe KB. Because propositional logic has limited
power of expression, as a future work, we will consider our
forgetting-based merging in a broad logic language such as
description logic which is proved to be a highly successful class of
knowledge representation languages in the Semantic Web.

\section*{Acknowledgements}
This work is supported by NSFC under grant number 60973003, 60496322
and the Ph.D. Programs Foundation of Ministry of Education of China.


\nocite{*}

\begin{thebibliography}{1}
\bibitem{liberatore1998arbitration}
Liberatore, P. and Schaerf, M., Arbitration (or how to merge
knowledge bases), IEEE Transactions on Knowledge and Data
Engineering, vol. 10(1), pp. 76--90, 1998.

\bibitem{konieczny2002merging}
Konieczny, S. and P{\'e}rez, R.P., Merging information under
constraints: a logical framework, Journal of Logic and Computation,
Oxford University Press, vol. 12, pp. 773--808, 2002.

\bibitem{gorogiannis2008merging}
Gorogiannis, N. and Hunter, A., Merging first-order knowledge using
dilation operators, in Proc. international conference on Foundations
of information and knowledge systems (FoIKS'08), ser. Lecture Notes
in Computer Science. Berlin, Germany: Springer, vol. 4932. pp.
132--150, 2008.

\bibitem{konieczny2004}
Konieczny, S., Lang, J. and Marquis, P., DA2 merging operators,
Artificial Intelligence, Elsevier, vol. 157(1-2), pp. 49--79, 2004.


\bibitem{dalal1988investigations}
Dalal, M., Investigations into a theory of knowledge base revision:
Preliminary report, in Proc. National Conference on Artificial
Intelligence (AAAI'88), AAAI Press / The MIT Press, vol. 2, pp.
475--479, 1988.

\bibitem{lin1999knowledge}
Lin, J. and Mendelzon, A.O., Knowledge base merging by majority,
Dynamic Worlds: From the Frame Problem to Knowledge Management, pp.
195-218, 1999.

\bibitem{revesz1993semantics}
Revesz, P.Z., On the semantics of theory change: arbitration between
old and new information, in Proc. {ACM} SIGACT-SIGMOD-SIGART
symposium on Principles of database systems (PODS'93), Washington,
DC. ACM Press pp.71--82, 1993.


\bibitem{silberschatz1990database}
Silberschatz, A., Stonebraker, M. and Ullman, J.D., Database
systems: Achievements and opportunities, Communications of the ACM,
Association for Computing Machinery, vol. 34(10), pp. 110--120,
1991.

\bibitem{Lin94forgetit!}
Lin,F. and Reiter,R., Forget It!, in Proc. {AAAI} Fall Symposium on
Relevance, pp. 154--159, 1994.

\bibitem{lang2010reasoning}
Lang, J. and Marquis, P., Reasoning under inconsistency: A
forgetting-based approach, Artificial Intelligence, Elsevier, vol.
174, pp. 799¨C823, 2010.




\end{thebibliography}
\end{document}